\documentclass[runningheads]{llncs}

\usepackage{eccv}

\usepackage{eccvabbrv}

\usepackage{graphicx}
\usepackage{booktabs}
\usepackage{algorithm}
\usepackage{algorithmic}

\usepackage[accsupp]{axessibility}  

\usepackage{hyperref}

\usepackage{orcidlink}

\begin{document}

\title{Inference-Time Attention Steering for Vision-Language-Action Driving Models} 

\titlerunning{Inference Time Attention Steering for VLA Models}

\author{Darshan Nagendra Prasad\inst{1} \and
Lars Ullrich\inst{2} \and
Knut Graichen\inst{3}}

\authorrunning{D.~Nagendra Prasad et al.}

\institute{FAU Erlangen-Nürnberg, Germany\\
\email{\{darshan.nagendra.prasad,lars.ullrich, knut.graichen\}@fau.de}}

\maketitle

\begin{abstract}
  Vision-language-action (VLA) driving models couple a reasoning stage with a diffusion-based trajectory decoder,
 but do not give a direct way to redirect attention toward safety-critical actors at inference time without retraining. 
  We studied a bounded additive pre-softmax attention bias on the visual tokens of detector localized traffic actors on Alpamayo-R1's Qwen3-VL backbone. It is applied as a fail open forward pre-hook with no weight changes. 
  On 50 lane-change scenarios from the PhysicalAI WorldModel-Synthetic dataset. The trajectory decoder shows 
  a monotonic dose response in the bias magnitude, separate from a paired zero bias control at every tested magnitude. 
  It reaches $\approx 17$\,cm mean displacement with lateral shifts up to $\sim 140$\,cm at the clamp. 
  A layer ablation places the action-relevant signal in late layers, where the effect increases with the number 
  of hooked layers (2.0\,cm for the first 8 layers; 67.6\,cm for all 36). A per call injection audit explains 
  why the Chain-of-Causation text never changes. The mask based bias never reaches the reasoning pathway in this 
  serving stack, so the invariance is verified \emph{exposure}, not robustness. Steered trajectories tend to 
  shift toward the attended actor, suggesting the bias governs where the model looks rather than encoding a 
  target behavior.
\end{abstract}

\section{Introduction}
\label{sec:intro}

Vision-language-action (VLA) models have emerged as a promising paradigm for end-to-end autonomous 
driving~\cite{wang2025alpamayo,hwang2024emma,tian2024drivevlm}. By jointly processing multi-camera inputs, 
generating natural-language reasoning chains, and producing trajectory waypoints, they unify perception, planning,
and control inside a single network. Alpamayo-R1~\cite{wang2025alpamayo} is built on a Qwen3-VL~\cite{qwen3vl2025} 
backbone with a diffusion-based action head. It is a state-of-the-art instance. It first emits a Chain-of-Causation 
(CoC) text describing the scene and the intended maneuver, and a diffusion sampler then decodes the future trajectory
conditioned on that reasoning and the visual context. Despite these advances, VLA models inherit a structural limitation: once deployed, their attention allocation over visual tokens is fixed by training. 
If the model under attends a safety-critical actor (a merging vehicle in a blind spot, a pedestrian at a crosswalk), 
there is no mechanism to redirect this at inference time without retraining. This is prohibitively expensive 
for 10B+ parameter models and breaks deployment modularity. Inference-time attention steering has been studied 
in the LLM literature~\cite{guardieiro2025instaboost,zhang2024pasta,venkateswaran2025spotlight} and risk-conditioned 
attention has been proposed for multimodal safety alignment~\cite{park2025moras}, but to our knowledge, no prior 
work has applied attention steering to a VLA driving model or studied whether such interventions propagate through 
hybrid architectures.

\begin{figure}[t] 
    \centering 
    \includegraphics[width=0.98\textwidth]{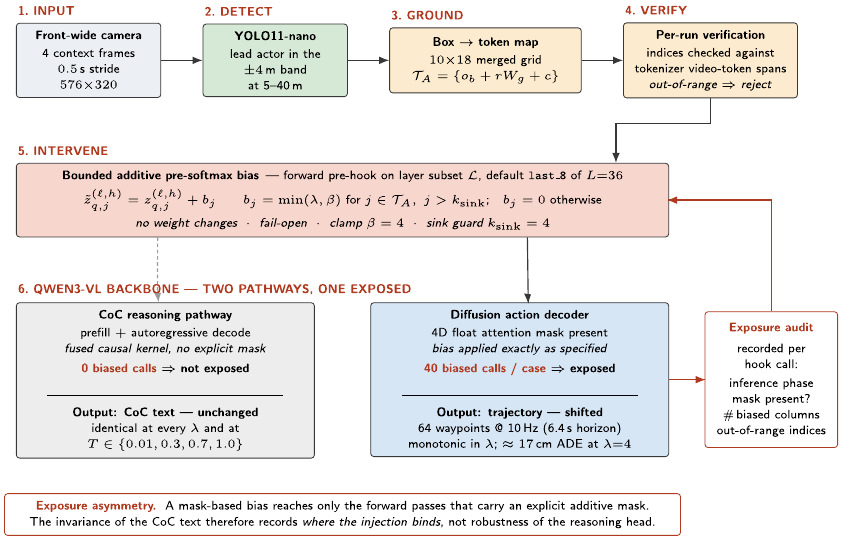} 
    \caption{Overview of the intervention. A detected lead actor is mapped 
    by \cref{eq:token_map} onto the merged visual-token grid. Every index is 
    verified against the tokenizer's video-token spans. And a forward pre-hook 
    adds the bounded bias of \cref{eq:bias_def} to those tokens' attention logits, 
    with no weight changes. The two pathways are not equally reachable: the diffusion 
    decoder carries an explicit 4D additive mask and receives the bias; the CoC path 
    uses a fused causal kernel with no mask and is never perturbed.} 
    \label{fig:pipeline} 
\end{figure}

\noindent\textbf{Approach.} 
We study a bounded additive pre-softmax attention bias targeting visual tokens of detector-localized traffic actors 
(\cref{fig:pipeline}). The bias is injected via lightweight forward pre-hooks on Qwen3-VL self-attention modules
with modifying the attention mask without altering any weights. The construction is: (a)~\emph{in\-fer\-ence-time}: no 
training or gradient computation; (b)~\emph{bounded}: clamped to prevent the suppression regime predicted by InstABoost 
theory~\cite{guardieiro2025instaboost}; (c)~\emph{modular}: the bias source is external and swapable; (d)~\emph{fail-open}: 
any failure reverts to native attention. A YOLO11-nano~\cite{ultralytics2024yolo11} detector run at the model's input resolution 
localizes a lead vehicle, whose bounding box is mapped through the patch grid and spatial merge factor of Qwen3-VL. It is mapped to the exact 
visual tokens of the input sequence, with the mapping verified per run against the tokenizer's video-token spans. The object of 
study is the \emph{propagation} of the intervention rather than its utility: whether, where, and by how much a bounded bias 
reaches the model's outputs. We make no claim that the resulting displacement constitutes better driving.

\noindent\textbf{Key findings.} 
On 50 lane-change scenarios from the PhysicalAI WorldModel-Synthetic dataset~\cite{nvidia2025wmsynth}, 
the diffusion trajectory decoder responds monotonically to the bias. Every steered case separates from 
the paired zero-bias control at every nonzero magnitude. A statement about detectability rather than magnitude, 
since the paired gate forces the control to be exactly zero. A layer ablation shows the effect concentrates in 
late transformer layers and accumulates with depth: the first 8 layers are nearly inert (2.0\,cm), while all 36 
layers exceed any contiguous subset (67.6\,cm). The CoC text never changes, but a per-call injection audit reveals why: 
in the deployed serving path, the additive-mask intervention binds exclusively to the action decoder's forward passes; 
the CoC prefill and decode steps invoke mask-free fused attention and are untouched by construction. The observed CoC invariance 
therefore reflects the realized exposure of the injection rather than architectural robustness. Inspection of the steered 
trajectories shows that, when the attended actor remains ahead of the ego, the trajectory tends to shift toward the actor's 
side. The magnitude grows with both the bias scale and the number of biased layers. We interpret this as evidence that the 
bias modulates the visual saliency of the actor in the representation consumed by the diffusion decoder, rather than encoding 
any explicit safety semantics.

\noindent\textbf{Contributions.} 
(i)~A systematic study of inference-time attention steering for a VLA autonomous driving model, 
with a detector-grounded actor-to-token pipeline verified per run. 
(ii)~A bounded pre-softmax bias formulation compatible with Qwen3-VL's grouped-query 
attention and per-head QK-Norm, implemented as a fail-open forward pre-hook. 
(iii)~A paired-seed validation methodology that isolates a centimeter-scale steering 
effect from a diffusion-sampler noise floor an order of magnitude larger, yielding a monotonic dose response. 
(iv)~A per-call exposure audit that corrects a tempting interpretation of CoC invariance as architectural r
obustness: the reasoning pathway was never perturbed in the deployed stack. We argue that apparent pathway selectivity of 
mask-based interventions on compound VLA models must be reported with audited exposure.

\section{Related Work} 
\label{sec:related} 

\noindent\textbf{Inference-time attention steering.} PASTA~\cite{zhang2024pasta} 
multiplicatively upweights post-softmax attention on user-specified tokens but requires materializing 
the attention matrix, breaking FlashAttention~\cite{dao2022flashattn}. InstABoost~\cite{guardieiro2025instaboost} 
applies a constant additive pre-softmax bias to instruction-key logits, showing that the bias tilts inter-rule 
competition exponentially while remaining kernel-compatible. SpotLight~\cite{venkateswaran2025spotlight} extends 
this with an additive bias that drives the attention proportion on a user-specified span toward a target, 
updating dynamically when the span is under-attended. Related interventions steer multimodal attention to 
reduce hallucination: VisFlow intervenes at the token and head level~\cite{tang2025visflow}, and ASCD~\cite{wang2026ascd} 
combines positive steering of text-centric heads with negative steering of visual tokens inside a contrastive-decoding loop. 
These methods all operate on text-output generation. Our intervention reuses the InstABoost additive-bias primitive but makes 
the bias \emph{scene-dynamic} (per actor, per frame) and \emph{actor-specific} (targeting detector-localized visual tokens 
of a specific real vehicle), and applies it to a driving model whose output is a physical trajectory rather than text. 

\noindent\textbf{Attention interventions in VLAs.} The closest work in the VLA domain uses attention interventions for 
\emph{diagnosis} rather than control. VLA-Trace~\cite{liu2026vlatrace} applies attention knockout (adding a large 
negative mask to selected attention logits before the softmax, with weights unchanged) to test whether specific token 
pathways are causally required for action generation. Our intervention is the sign flipped counterpart, we add a 
bounded positive bias to amplify a pathway and measure the resulting trajectory change rather than ablating a pathway 
to test necessity. Gaze-Regularized VLA~\cite{pani2026gaze} instead shapes attention at \emph{training time}, 
adding a KL term that aligns the policy's vision-language attention with human gaze. This would change weights and incurs 
no inference-time intervention, whereas our method modifies no weights and is acting only at inference. 

\noindent\textbf{Risk-aware driving.} Risk Semantic Distillation~\cite{qin2025rsd} 
distills VLM-derived risk semantics into BEV features at \emph{training time}. MoRAS~\cite{park2025moras} proposes risk-adaptive 
activation steering for multimodal safety alignment. Identifying the insufficient visual attention to safety-critical regions as a 
key failure mode. Our work differs from RSD in operating at inference time without retraining, and from MoRAS in targeting an 
autonomous driving model with a diffusion action head rather than jailbreak defense on a text decoder. 

\noindent\textbf{VLA interpretability and architectural priors.} Recent work shows that multimodal attention 
flow is layer-dependent, with late layers carrying the bulk of action-relevant signal~\cite{yang2024dleaf}, 
and that LLMs concentrate disproportionate attention mass on initial-token ``sinks'' whose disruption destabilizes 
generation~\cite{xiao2024sinks}. These motivate our layer ablation and our sink-guarding design, in which the first 
$k_{\text{sink}}$ positions are excluded from bias injection.

\section{Method} 
\label{sec:method} 
\subsection{Bounded Additive Pre-Softmax Bias} 
\label{sec:bias} 
Let the VLA model $\mathcal{M}$ process a token sequence 
$\mathbf{x} = [x_1, \ldots, x_N]$ of interleaved text and visual tokens. 
Given an actor $A$ with associated visual token set $\mathcal{T}_A \subset \{1,\ldots,N\}$, 
we wish to increase the model's attention mass on $\mathcal{T}_A$ \emph{without modifying any weights}.

For Qwen3-VL's grouped-query attention with per-head QK-Norm~\cite{qwen3vl2025}, 
the standard attention logit between query position $q$ and key position $j$ at layer $\ell$, head group $h$ is

\begin{equation}
    \label{eq:attn_logit}
    z_{q,j}^{(\ell,h)} = \frac{\mathbf{q}_q^{(\ell,h)} \cdot \mathbf{k}_j^{(\ell,h)}}{\sqrt{d_h}},
\end{equation}
with softmax attention weights
$\alpha_{q,j}^{(\ell,h)} = \exp(z_{q,j}^{(\ell,h)}) /
\sum_{j'} \exp(z_{q,j'}^{(\ell,h)})$.
Our intervention adds a bounded bias vector $\mathbf{b}\in\mathbb{R}^N$:
\begin{equation} 
    \label{eq:bias_def} 
    \tilde{z}_{q,j}^{(\ell,h)} = z_{q,j}^{(\ell,h)} + b_j, \quad 
    b_j = \begin{cases} 
        \min(\lambda,\beta) & j \in \mathcal{T}_A,\ j > k_{\text{sink}}, \\ 
        0 & \text{otherwise}, 
    \end{cases} 
\end{equation}
where $\lambda \geq 0$ is the \emph{head\_scale} parameter controlling bias magnitude, 
$\beta > 0$ a safety clamp, and $k_{\text{sink}}$ the number of sink-guarded initial positions.

\begin{proposition}[Attention mass on the targeted tokens]
\label{prop:monotone}
Fix a query position $q$, layer $\ell$, and head group $h$, and let
$m_{\ell,h,q}(A) = \sum_{j\in\mathcal{T}_A} \alpha_{q,j}^{(\ell,h)}$ denote the
attention mass placed on the targeted tokens. For $0 < \lambda \leq \beta$, the
biased mass $\tilde{m}$ is at least $m$ and increases with $\lambda$, with
equality only in degenerate cases (\eg, $\mathcal{T}_A$ already carrying all
mass).
\end{proposition}

This follows from standard softmax algebra and is the additive form of InstABoost's 
Theorem~1~\cite{guardieiro2025instaboost}: adding $\lambda$ to the logits of $\mathcal{T}_A$ 
scales their softmax numerators by $e^{\lambda}>1$ and leaves the others unchanged. Then the 
renormalized weights on $\mathcal{T}_A$ cannot decrease. We are mentioning it here itself to make explicit 
the direction of the intervention. We do not claim it predicts the resulting trajectory change, 
which is the empirical question studied in \cref{sec:results}. One practical consequence is that 
any $\lambda \geq \beta$ applies the same effective bias $\beta$; consistent with this, we observe 
bit-identical trajectories at $\lambda\in\{4,8,10\}$ for $\beta=4$.

The clamp $\beta$ prevents the \emph{suppression regime}: 
excessively large biases cause instruction over-focus that degrades 
generation~\cite{guardieiro2025instaboost}. The additive pre-softmax form 
is compatible with FlashAttention~\cite{dao2022flashattn,press2022alibi,wu2025flashbias} 
. Unlike key-vector editing it preserves the learned scale calibration of QK products 
under per-head QK-Norm, since the bias is applied \emph{after} the norm-stabilized dot product.

\subsection{Actor Detection and Token Mapping} 
\label{sec:mapping} 
A YOLO11-nano detector~\cite{ultralytics2024yolo11} runs at the model's input 
resolution ($W{\times}H = 576{\times}320$) on the most recent context frame of the 
front camera. The detector and the vision encoder share coordinates and makes sure that no rescaling 
error can enter the mapping. Box $(x_1,y_1,x_2,y_2)$ yields a pinhole range estimate $d = f_y H_A / h_{\text{px}}$ 
from a class-prior height $H_A$, and a lateral offset from the box centroid. The \emph{lead actor} is the nearest 
vehicle within a $\pm 4$~m lateral band (covering the ego and adjacent lanes) at 5--40~m longitudinal range. If no 
detection satisfies these constraints, the scenario is rejected upstream. The detector is not part of the contribution, 
the intervention's only interface is a box in model-input coordinates. Any detector producing one substitutes without 
touching the hook or the mapping. We chose the smallest YOLO11 variant because it runs at the model's native resolution, 
removing rescaling as a source of mapping error.

\begin{figure}[t] 
    \centering 
    \includegraphics[width=0.85\textwidth]{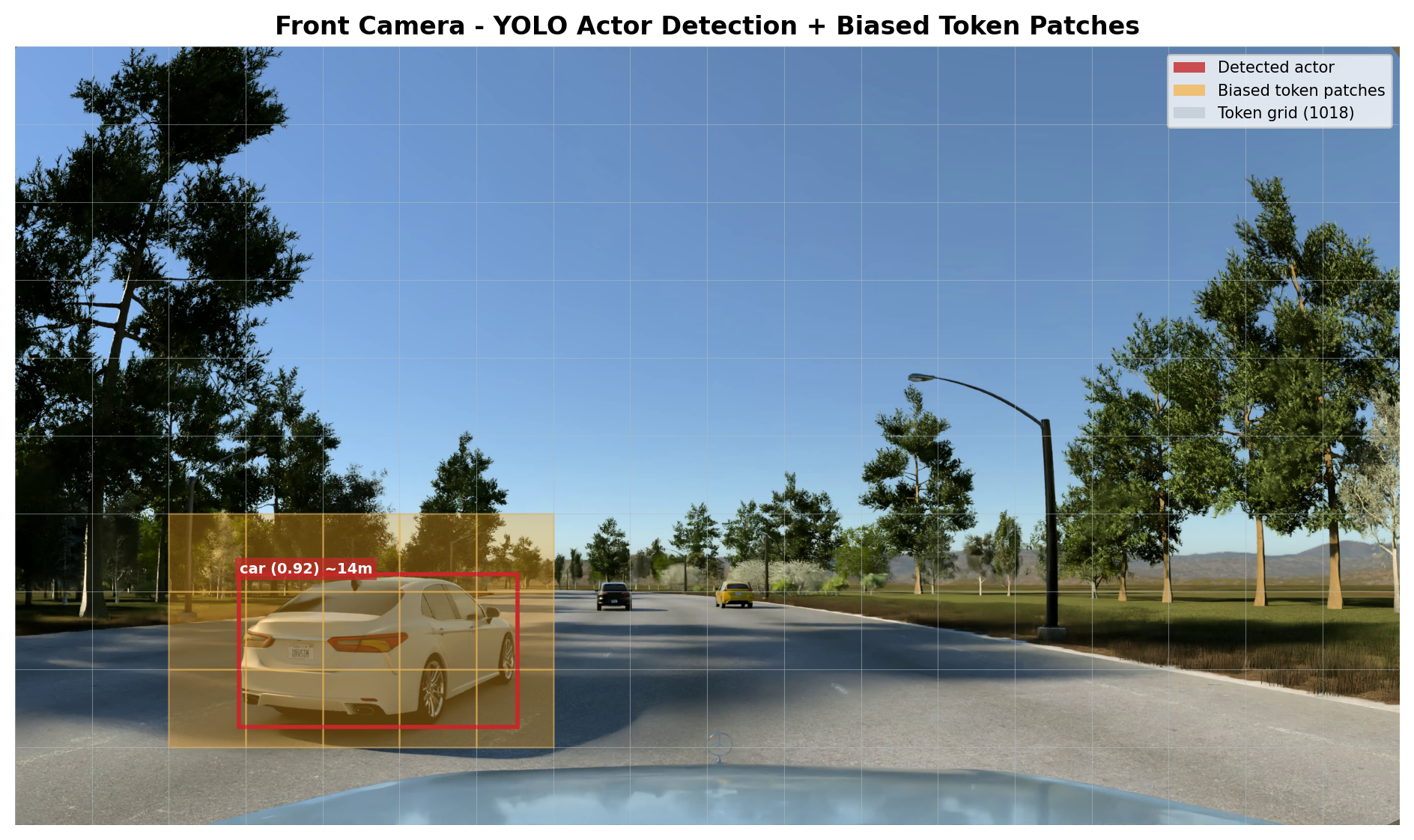} 
    \caption{Front-camera actor grounding. The YOLO11 detector localizes the lead vehicle 
    (\emph{red box}, with confidence and pinhole range estimate); the box is mapped through 
    \cref{eq:token_map} onto the $10{\times}18$ merged token grid (\emph{thin overlay}); 
    the resulting token patch (\emph{orange cells}) receives the pre-softmax bias. 
    The mapping is verified per run against the tokenizer's video-token spans, so any 
    resolution mismatch is detected before injection.} 
    \label{fig:overlay} 
\end{figure}

Qwen3-VL's vision encoder compresses each frame through a patch size $p$ and a 
$2{\times}2$ spatial merger~\cite{qwen3vl2025}, producing a token grid 
\begin{equation} 
    \label{eq:grid} 
    (H_g, W_g) = (H/(pm),\ W/(pm)) = (10, 18) 
\end{equation} 
for our resolution, read directly from the model's \texttt{image\_grid\_thw} 
output rather than assumed. The full token sequence interleaves one such block per (camera, frame) pair. 
Mapping a box to tokens proceeds by cell arithmetic: 
\begin{equation} 
    \label{eq:token_map} 
    \mathcal{T}_A = \{o_b + rW_g + c \mid r\in[r_1, r_2]\cap[0,H_g),\ c\in[c_1, c_2]\cap[0,W_g)\}, 
\end{equation} 
with $r_i = \lfloor y_i/(pm) \rfloor$, $c_i = \lfloor x_i/(pm) \rfloor$, and 
$o_b$ the sequence offset of the corresponding (camera, frame) block. Every produced index is 
\emph{verified} against the tokenizer's video-token spans; any run with out-of-range targets is rejected, as illustrated in 
\cref{fig:overlay}. This verification matters because subtle changes in the processor's resize pipeline can shift 
the effective resolution and silently invalidate the mapping otherwise.

\subsection{Implementation and Exposure Audit}
\label{sec:hooks}

The bias is implemented through \texttt{register\_forward\_pre\_hook} on each
target \texttt{Qwen3VL\allowbreak Text\allowbreak Attention} module
(\cref{alg:hook}). The hook edits the \texttt{attention\_\allowbreak mask}
keyword argument: in Qwen3-VL's eager
attention path. This is a 4D float tensor of shape $[B,1,L_q,L_k]$ with masked
positions at $-\infty$ and unmasked at $0$, so adding $+\lambda$ to columns
$j\in\mathcal{T}_A$ is algebraically equivalent to \cref{eq:bias_def}. The hook
is fail open, if the keyword argument is absent or contains unexpected shapes,
the inputs are returned unchanged and the skip is logged.

\begin{algorithm}[t]
\caption{Bias Injection via Forward Pre-Hook with Exposure Audit}
\label{alg:hook}
\begin{algorithmic}[1]
\REQUIRE Layer module $\ell$, target indices $\mathcal{T}_A$, scale $\lambda$, clamp $\beta$, sink guard $k_{\text{sink}}$
\STATE $\mathcal{T}_A' \leftarrow \{j\in\mathcal{T}_A : j > k_{\text{sink}}\}$
\IF{$\mathcal{T}_A' = \emptyset$ \textbf{or} \texttt{attention\_mask} absent}
    \STATE \textit{record skip (phase, reason)}; \textbf{return} unmodified inputs
\ENDIF
\STATE $\mathbf{b}\leftarrow\mathbf{0}_{L_k}$;\ \ \textbf{for} $j\in\mathcal{T}_A'$ \textbf{do} $b_j \leftarrow \min(\lambda,\beta)$
\STATE $\texttt{attention\_mask} \leftarrow \texttt{attention\_mask} + \mathbf{b}$
\STATE \textit{record application (phase, $|\mathcal{T}_A'|$, effective bias)}
\STATE \textbf{return} modified inputs
\end{algorithmic}
\end{algorithm}

\noindent\textbf{Exposure audit.} A mask based injection can only act where an
explicit additive mask exists. In the deployed Alpamayo-R1 model serving path, the
CoC prefill and autoregressive decode invoke attention through the fused causal
kernel \emph{without} an explicit mask. While in the diffusion action decoders 
it forward passes have a 4D float mask. The injector therefore checks each hook 
call noting the inference phase, mask presence, number of columns and any out of range indices.
This audit defines the intervention's
\emph{exposure} (the set of forward passes it actually perturbs) and is
essential for interpreting the CoC results in \cref{sec:exposure}.

\noindent\textbf{Layer selection.} We define a layer subset
$\mathcal{L}\subseteq\{0,\ldots,L-1\}$ where hooks are installed:
\texttt{all}, \texttt{last\_k}, \texttt{first\_k}, or \texttt{mid\_k}; ablated
in \cref{sec:layer}.

\section{Experimental Setup}
\label{sec:setup}

\noindent\textbf{Model.} Alpamayo-R1-10B~\cite{wang2025alpamayo} loaded with
NF4 quantization~\cite{dettmers2024qlora} ($\approx 12$\,GB VRAM, single 24\,GB
GPU). The Qwen3-VL backbone has $L=36$ transformer layers with grouped-query
attention (8 KV heads, 28 query heads). Trajectories comprise 64 waypoints at
10\,Hz (6.4\,s horizon).

\noindent\textbf{Data.} 50 lane-change scenarios from the PhysicalAI
WorldModel-Synthetic Autonomous Driving Scenarios dataset~\cite{nvidia2025wmsynth},
comprising Omniverse-rendered multi-camera videos. We use the front-wide camera
with four context frames at 0.5\,s stride, resized to $576{\times}320$. Cases
are selected by a ranked procedure requiring a detected lead vehicle in the
$\pm 4$\,m band at 5--40\,m range, preferring $\sim 20$\,m, and deduplicated
against the dataset's environment-permutation expansion of authored scenarios.
The synthetic setting is a deliberate control choice, not a fallback: detector
variance is indistinguishable, in the trajectory delta, from bias variance. On
recorded logs the lead-actor definition degrades in exactly the ways that would
confound this parked vehicles satisfy a range-and-lane test while playing no
role in the maneuver, turns carry the actor out of frame within the horizon, and
truncated or occluded vehicles yield boxes whose mapped token set does not cover
the actor. Curating around these failure modes is a perception problem
orthogonal to the question studied here. Authored lane-change scenarios give one
unambiguous lead actor held in frame across the context window; the cost is
external validity (\cref{sec:discussion}).
The published videos do not include ego state; we synthesize a constant-velocity,
straight-line ego history (10\,m/s) matching the tensor conventions of a recorded
reference clip. Two safeguards address the resulting distribution shift: the
baseline CoC must be a coherent driving description, and a \emph{zero-bias gate}
requires head\_scale${}=0$ to reproduce the baseline trajectory exactly under
paired seeds (ADE $\equiv 0$); all admitted cases pass both.

\noindent\textbf{Paired-seed protocol.} Diffusion trajectory sampling is
stochastic. We measure all effects \emph{paired}: baseline and steered runs
share the identical seed and deterministic-algorithms configuration, so any
nonzero difference is attributable to the bias. The zero-bias gate verifies the
pairing is exact (ADE = 0 under bit-identical seeds).

\noindent\textbf{Configuration.} Decoding temperature $T = 0.01$ for CoC (varied
in \cref{sec:exposure}). Bias: \texttt{max\_\allowbreak abs\_\allowbreak bias} $\beta=4.0$,
\texttt{sink\_\allowbreak guard\_\allowbreak tokens} $k_{\text{sink}}=4$;
default \texttt{layer\_\allowbreak subset}${}={}$\texttt{last\_8}, ablated in
\cref{sec:layer}.

\noindent\textbf{Metrics.} (i) Average Displacement Error (ADE): mean $L_2$
distance between paired baseline and steered 64-waypoint trajectories.
(ii) Maximum lateral shift over the horizon.
(iii) CoC change rate under exact string match.
(iv) Cliff's $\delta$~\cite{cliff1993dominance} against the paired zero-bias
control.
(v) Wilcoxon signed-rank tests~\cite{wilcoxon1945individual} with Holm
correction~\cite{holm1979simple}.
(vi) Shift direction relative to the attended actor's side.
These quantify how far and in which direction the intervention moves the
prediction. None is a driving-quality score, and we make no claim that a larger
displacement is a better one. The
direction analysis is only computed in the subset of timesteps where the
actor remains ahead of the ego under a constant-velocity actor model; we
report the actor-ahead fraction alongside the directional fractions and treat
the analysis as informative only when this fraction exceeds a per-clip
threshold.

\section{Results}
\label{sec:results}

\subsection{Injection Verification and Sampler Noise Floor}
\label{sec:verify}

For 5 cases, we capture each hooked layer's attention mask before and after
bias installation at head\_scale${}=4.0$. The mean logit delta equals $+4.0$
at all target columns and $0.0$ at all non-target columns, with zero
out-of-range target events; the audit confirms 40 biased attention calls per
case, exclusively during the diffusion action decoder's forward passes
(\cref{sec:exposure} expands on the calls that are \emph{not} biased). The
intervention is therefore implemented exactly as specified by
\cref{eq:bias_def}.

To separate steering from sampling stochasticity, we run 3 cases at 5 seeds
each. Same-seed repeats are bit-identical (intra-seed ADE $=0$, confirming the
zero-bias gate). The seed-paired steering effect at head\_scale${}=4$ averages
$0.107$\,m across these cases, while the cross-seed
baseline-to-baseline distance averages $3.23$\,m, a factor $\approx 30$
\emph{larger} than the effect. The paired protocol is therefore necessary:
an unpaired comparison would bury the steering effect entirely in sampler
variance. The diffusion noise floor reflects the genuine stochasticity of the
trajectory sampler at non-zero temperature in the latent space; the bias-induced
shift is a much smaller, but reproducible, perturbation of the same sampling
distribution.

\subsection{Dose--Response and Statistical Significance}
\label{sec:dose}

\begin{figure}[t]
\centering
\includegraphics[width=\textwidth]{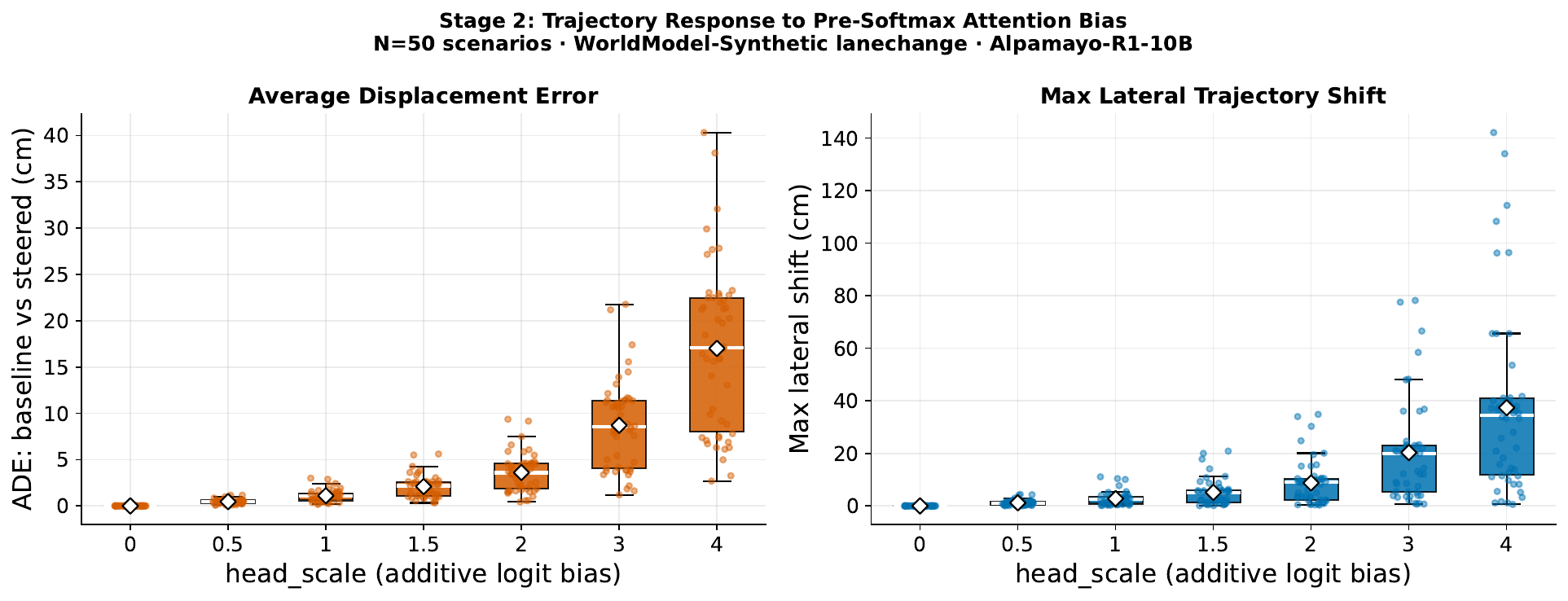}
\caption{Trajectory response to pre-softmax attention bias on detector-grounded
actor tokens, across 50 lane-change scenarios. Boxes show the IQR; the white
line is the median, the diamond the mean, and individual points are
per-scenario values. \emph{Left:} ADE grows monotonically with bias magnitude,
reaching $\approx 17$\,cm mean at head\_scale${}=4$. \emph{Right:} maximum
lateral shift reaches $\approx 38$\,cm mean, with a heavy upper tail (single
cases up to $\sim 140$\,cm). The zero-bias condition is exactly zero in every
case by construction of the paired-seed gate.}
\label{fig:boxstrip}
\end{figure}

\Cref{fig:boxstrip} reports the head\_scale sweep over
$\{0, 0.5, 1, 1.5, 2, 3, 4\}$ for all 50 cases (the sweep stops at $\beta = 4$
since larger values are clamped duplicates of $\beta$ by construction).

ADE grows monotonically from sub-centimeter values at $\lambda{=}0.5$ to
$\approx 17$\,cm mean at $\lambda{=}4$, while the zero-bias condition is
exactly zero in every case. All six nonzero conditions are highly significant
against the paired zero-bias control after Holm correction
($p \ll 10^{-10}$) with Cliff's $\delta = 1.0$ throughout: \emph{every}
steered case exceeds \emph{every} zero-bias case at every magnitude. The
per-case spread widens with $\lambda$, indicating scenario-dependent
susceptibility; the largest single-case lateral shift reaches $\sim 140$\,cm
at saturation. Effects are predominantly longitudinal at small $\lambda$,
with the lateral component growing with bias magnitude until it dominates
the ADE in most cases at $\lambda{=}4$.

\noindent\textbf{Cliff's $\boldsymbol{\delta=1.0}$ versus dose--response: what
the data say and do not say.} The Cliff's-$\delta$ result is a statement about
\emph{detectability}, not about magnitude. Because the paired zero-bias gate
forces the baseline ADE distribution to be a Dirac mass at zero, any nonzero
steered ADE is strictly larger and contributes to $\delta = 1.0$. The
dose response figure conveys the magnitude information that $\delta$ does not:
the median ADE at $\lambda{=}0.5$ is below 1\,cm, with several scenarios
producing near-zero shifts even at intermediate $\lambda$. The combined reading
is that the bias has a \emph{consistently detectable} effect at every nonzero
magnitude, while the operationally relevant effect (one that visibly shifts the
trajectory) requires moderate to large $\lambda$ and is heterogeneous across
scenarios. Because $\delta=1.0$ follows largely from how the comparison is
constructed, we report it only as a consistency check on the paired protocol
and rest no further claim on it.

\subsection{Layer Localization}
\label{sec:layer}

\begin{figure}[t]
    \centering
    \includegraphics[width=\textwidth]{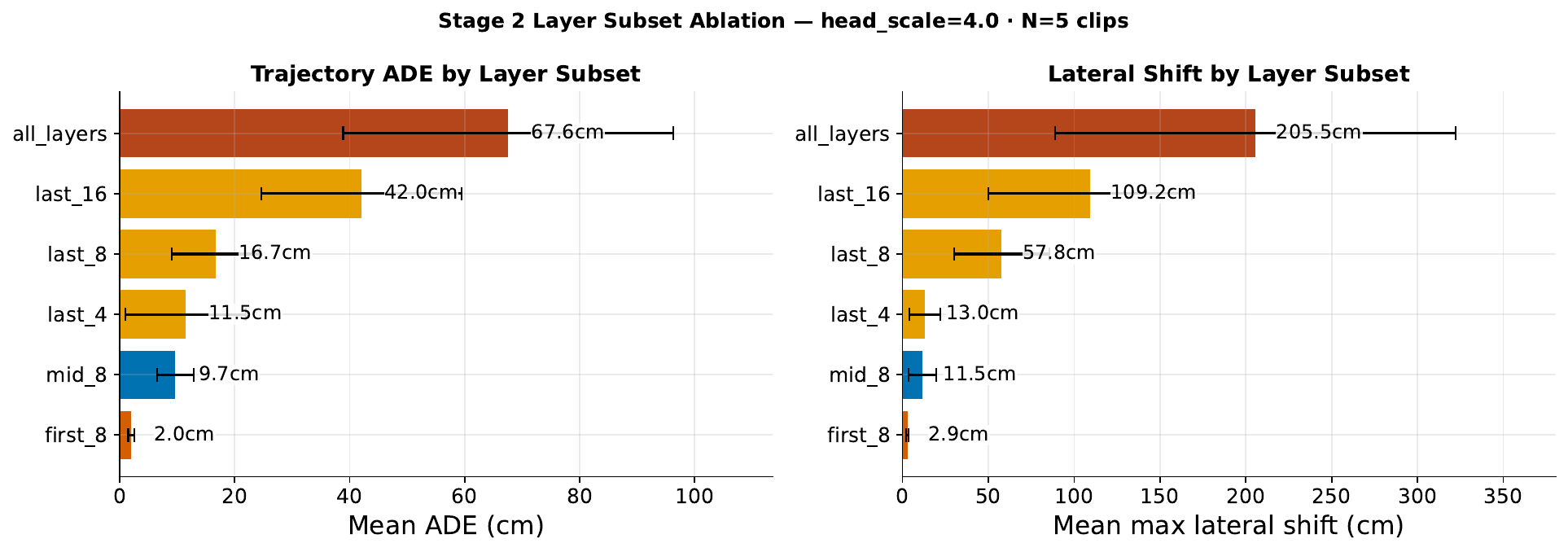}
    \caption{Layer subset ablation at head\_scale${}=4.0$ on detector-grounded
    actor tokens ($N{=}5$ clips). Early layers (\texttt{first\_8}) are nearly
    inert; late layers dominate; the effect accumulates with the number of hooked
    layers, and \texttt{all\_layers} exceeds every contiguous subset on both
    metrics.}
    \label{fig:ablation}
\end{figure}

\Cref{fig:ablation} reports six layer subsets at head\_scale${}=4.0$ over 5
cases. Mean ADE: \texttt{first\_8} 2.0\,cm, \texttt{mid\_8} 9.7\,cm,
\texttt{last\_4} 11.5\,cm, \texttt{last\_8} 16.7\,cm, \texttt{last\_16}
42.0\,cm, \texttt{all\_layers} 67.6\,cm. The corresponding maximum lateral
shifts follow the same ordering, with \texttt{all\_layers} averaging
$\sim 2$\,m across the five clips (\cref{tab:layer_ablation}).

\begin{table}[t]
    \centering
    \caption{Layer-subset ablation. Mean ADE and mean max lateral shift, at
    head\_scale${}=4.0$ across $N=5$ clips. Hooked count refers to the number of
    transformer layers in which the pre-hook is installed.}
    \label{tab:layer_ablation}
    \begin{tabular}{lccc}
        \toprule
        Layer subset    & Hooked layers & Mean ADE (cm) & Mean max lat.\ shift (cm) \\
        \midrule
        \texttt{first\_8}   & 8  & 2.0  & 2.9   \\
        \texttt{mid\_8}     & 8  & 9.7  & 11.5  \\
        \texttt{last\_4}    & 4  & 11.5 & 13.0  \\
        \texttt{last\_8}    & 8  & 16.7 & 57.8  \\
        \texttt{last\_16}   & 16 & 42.0 & 109.2 \\
        \texttt{all\_layers}& 36 & 67.6 & 205.5 \\
        \bottomrule
    \end{tabular}
\end{table}

Three structural observations follow. First, early layers are nearly inert, 
biasing only the first 8 transformer layers produces a 2.0\,cm mean ADE,
roughly $30\times$ below the all-layer response and close to the operational
floor of the effect.
Second, late layers dominate: \texttt{last\_8} (16.7\,cm) substantially exceeds
\texttt{mid\_8} (9.7\,cm) and \texttt{last\_4} (11.5\,cm) at the same
hooked-layer count. It is consistent with predictions from the multimodal attention
literature that late layers carry the bulk of the action-relevant
signal~\cite{yang2024dleaf}. 
Third, the effect accumulates roughly monotonically with the number of hooked layers within the late-layer regime:
\texttt{last\_4} $<$ \texttt{last\_8} $<$ \texttt{last\_16} $<$ \texttt{all},
on both ADE and max lateral shift. We interpret this depth accumulating
pattern as a useful empirical knob: the magnitude of the trajectory response
can be modulated either by the bias scale $\lambda$ or by the layer subset
$\mathcal{L}$, with the latter giving finer-grained control at fixed $\lambda$.

\subsection{Trajectory Geometry and Direction}
\label{sec:direction}

\begin{figure}[t]
    \centering
    \includegraphics[width=\textwidth]{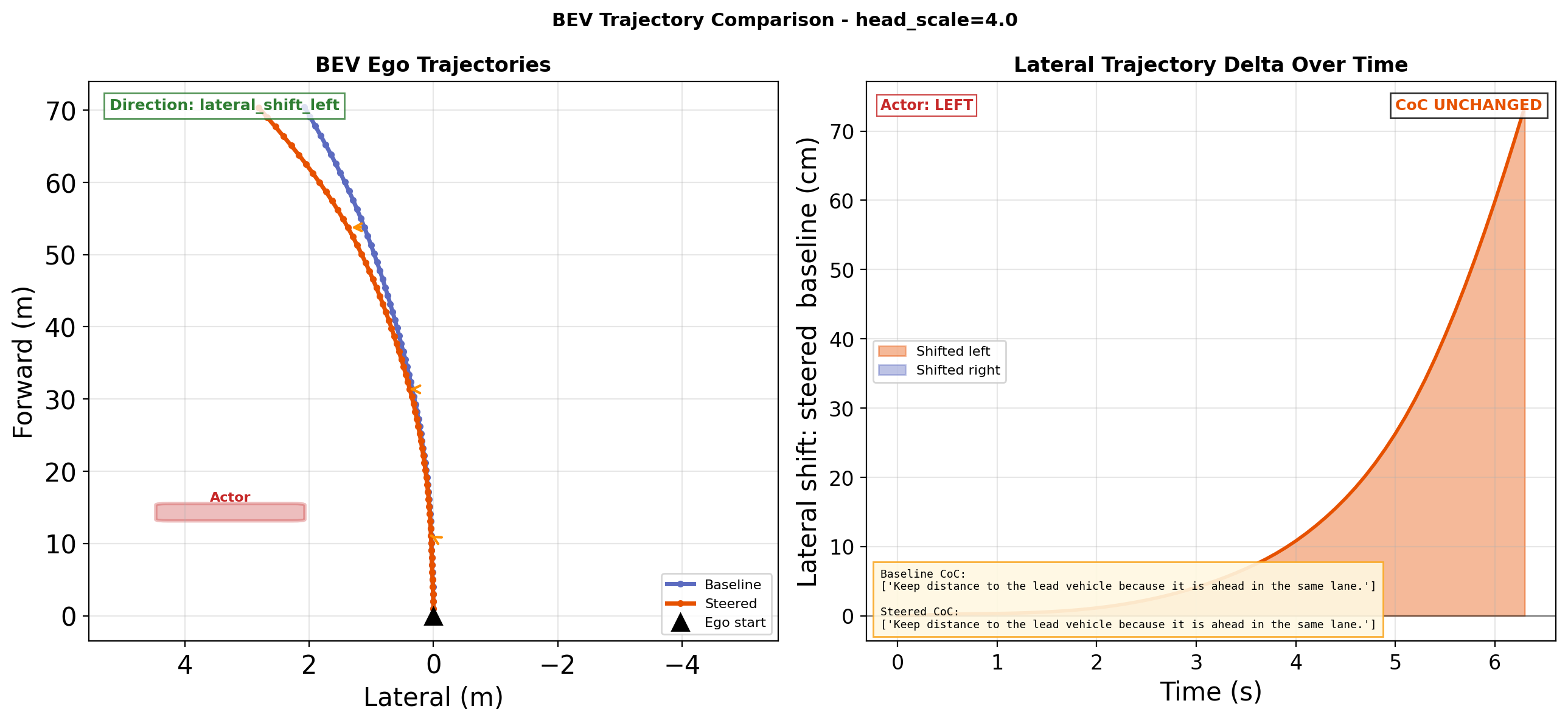}
    \caption{Bird's-eye-view paired comparison at head\_scale${}=4.0$. \emph{Left:}
    baseline (blue) and steered (orange) ego trajectories over a 6.4\,s horizon;
    the detected actor is shown in red on the ego's left. \emph{Right:} lateral
    delta (steered $-$ baseline) over time. The steered trajectory deviates
    progressively toward the actor's side as the rollout proceeds. CoC text is
    identical between the two runs.}
    \label{fig:bev}
\end{figure}

The summary statistics in \cref{sec:dose,sec:layer} convey magnitude but not
geometry. \Cref{fig:bev} shows a representative paired comparison at
$\lambda=4$. The steered ego trajectory tracks the baseline closely for the
first $\sim 2$\,s of the rollout and then deviates progressively toward the
actor's side; the lateral component grows roughly quadratically in the
deviation phase and reaches $\sim 74$\,cm at the horizon, while the
longitudinal component remains small. The CoC text is identical between
baseline and steered runs, as it is in every paired comparison in this study
(see \cref{sec:exposure}).

\begin{figure}[t]
    \centering
    \includegraphics[width=\textwidth]{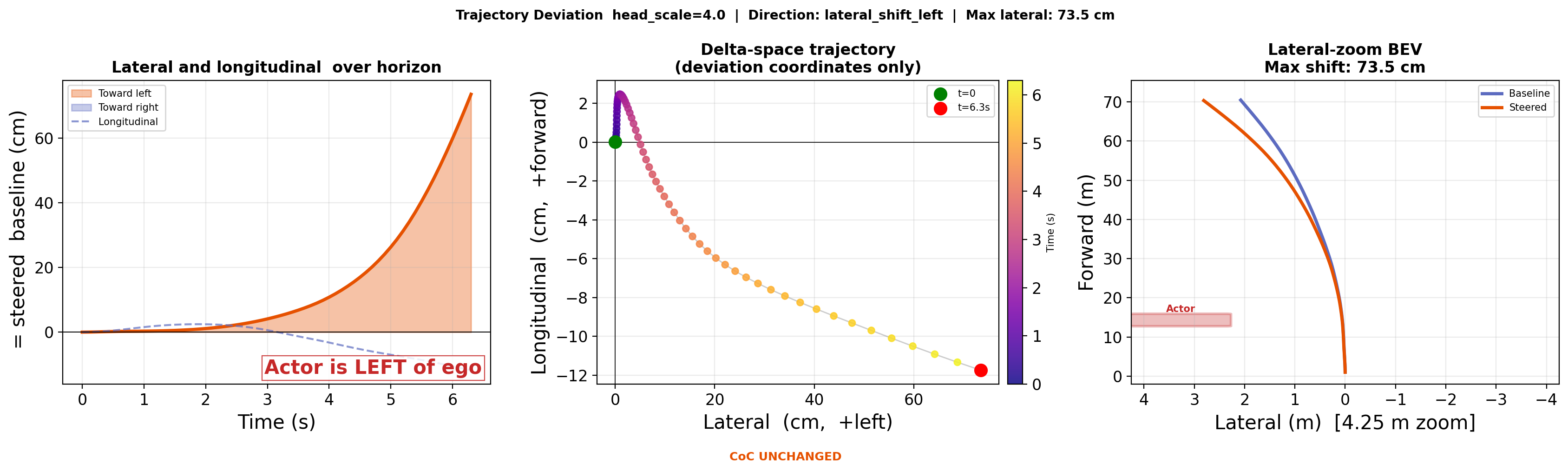}
    \caption{Trajectory-deviation panel for the same clip. \emph{Left:} lateral
    delta over time (filled) and longitudinal delta (dashed). \emph{Middle:}
    deviation in (lateral, longitudinal) coordinates with time coloring. \emph{Right:}
    lateral-zoom BEV emphasizing the small but reproducible lateral offset relative
    to the baseline. Maximum lateral shift is 73.5\,cm at $t \approx 6.3$\,s.}
    \label{fig:delta}
\end{figure}

\noindent\textbf{Toward-actor shift: a careful reading.} A formal towards/away
direction analysis projects the per-timestep steered--baseline displacement
onto the unit vector from each baseline waypoint to the actor (under a
constant-velocity actor model), and is only meaningful while the actor remains
ahead of the ego. Across the scenarios in which this validity criterion is
satisfied for a substantial fraction of the horizon, qualitative inspection of
the paired trajectories indicates that the steered trajectory shifts
predominantly toward the side on which the attended actor sits, rather than
away from it. We avoid a stronger claim: the relevant counterfactual
(``the model attended the actor and chose to drive closer to it'') is not
directly testable from open-loop trajectories, and a fraction of
the per-timestep direction signal flips sign depending on actor lateral position
relative to the ego heading. We draw only the following operational
reading. The bias appears to act as an \emph{attractor} on the diffusion
decoder's conditioning, increasing the actor's effective saliency in the
aggregate visual representation; whether this should be interpreted as a
safety-positive or safety-negative effect depends on the actor's role (a lead
vehicle to follow versus an obstacle to avoid). Mapping a desired behavior
(\eg, increased clearance to a hazard versus tighter lane-following behind a
lead) onto bias sign, magnitude, and target token set is therefore an open
design problem that the present framework can now evaluate quantitatively.

\noindent\textbf{Progressive divergence.} The lateral delta in
\cref{fig:bev,fig:delta} is near zero in the first 2\,s and grows
super-linearly thereafter, reaching its peak near the end of the horizon. We
observe this rollout-accumulating pattern broadly across the dose--response
set: the early-horizon shift is small relative to the late-horizon shift,
consistent with the diffusion decoder integrating the perturbed visual
conditioning over the predicted waypoint sequence rather than responding to it
in a single step. A consequence for evaluation is that ADE understates the
operationally important late-horizon shift; max lateral shift is the more
faithful summary in this regime.

\subsection{Chain-of-Causation Invariance: Exposure, Not Robustness}
\label{sec:exposure}

The CoC text is identical in all paired comparisons across the
dose--response sweep, at every bias magnitude, and across four CoC decoding
temperatures, $T \in \{0.01, 0.3, 0.7, 1.0\}$, on a five-case subset. A natural reading is that the autoregressive head is robust to the
attention perturbation while the diffusion head is sensitive, a clean
architectural asymmetry.

The per-call exposure audit shows that this reading would be incorrect. In the
deployed serving configuration, the count of biased CoC prefill calls and
biased autoregressive decode calls is \emph{zero} in every run. The CoC
prefill/decode path invokes attention through the fused causal kernel
\emph{without} an explicit additive mask; the additive-mask intervention
therefore never reaches it. Two conclusions follow. First, the temperature
sweep rules out deterministic decoding as a trivial explanation for invariance
\emph{given the realized exposure}: even at $T=1.0$ the CoC is unchanged
because the bias is not applied to its forward passes. Second, and more
importantly, the invariance we observe is an \emph{attribution} result rather
than a robustness property: the perturbation did not touch the reasoning
pathway, so whether the autoregressive head would be robust to an attention
perturbation that actually reaches it remains open. Establishing language-path
exposure, or proving it unreachable by mask-based means in this stack, is the
most important follow-up enabled by the audit methodology, and we consider it
out of scope for the present study.

\section{Discussion and Limitations}
\label{sec:discussion}

\noindent\textbf{From ``architectural asymmetry'' to verified exposure.}
What the data establish is not a difference in how the two heads \emph{process} the perturbation, but a precise and reproducible \emph{exposure asymmetry} of mask-based interventions. Reporting audited exposure alongside effects should therefore be a default for intervention studies on compound multimodal architectures, since apparent pathway selectivity may otherwise be an artifact of where the injection mechanism binds.

\noindent\textbf{Implications for safety-aware VLA design.}
The observed exposure asymmetry has dual implications. On the positive side, a
risk module can modulate motion planning without disturbing the interpretable
reasoning chain, and because exposure is auditable it can \emph{prove} that
the reasoning trace was not manipulated. On the negative side, the CoC text
cannot serve as a monitor for trajectory-level interventions, since the
reasoning pathway is blind to a bias that demonstrably moves the executed
trajectory by tens of centimeters. The toward-actor tendency observed in
\cref{sec:direction} further suggests that mapping a desired behavior onto bias
sign, magnitude, and placement is itself a design problem; composing the
present primitive with formal safety envelopes such as RSS~\cite{shalev2017rss}
is an open question.

\noindent\textbf{Scope of evidence.} The dose--response, layer ablation,
exposure audit, and direction analysis together establish that (a) the
intervention is implemented exactly as specified, (b) it produces a monotonic,
paired-significant trajectory response that grows with both bias magnitude and
the number of biased late-layer modules, (c) its effects accumulate over the
rollout horizon, and (d) the observed CoC invariance reflects verified
non-exposure rather than head-level robustness. The evidence is consistent
with the interpretation that the bias modulates visual saliency in the
representation consumed by the diffusion decoder. We stop short
of stronger causal claims about the model's high-level behavior.

\noindent\textbf{Limitations.}
\emph{(i) Synthetic environment.} The controlled setting that makes the
actor-grounding unambiguous (\cref{sec:setup}) also bounds the claim: the videos
are Omniverse-rendered~\cite{nvidia2025wmsynth} without recorded ego state, and
the ego history is synthesized. Both safeguards (coherent baseline CoC; exact
zero-bias gate) pass in all admitted cases, but sim-to-real transfer of the
magnitudes is untested, and whether the dose--response survives the messier
actor geometry of recorded logs is open.
\emph{(ii) Open-loop evaluation.} Trajectories are evaluated as predictions
of a single forward pass; closed-loop metrics on benchmarks such as
Bench2Drive~\cite{jia2024bench2drive} require simulation infrastructure beyond
this study.
\emph{(iii) Single model and lane-change focus.} All experiments use
Alpamayo-R1-10B on lane-change scenarios; generalization to other VLAs and
interaction types (pedestrians, intersections, multi-actor scenes) is
hypothesized but unverified.
\emph{(iv) Lan\-guage-path exposure not established.} As shown in
\cref{sec:exposure}, the reasoning pathway was never actually perturbed by the
mask-based injection in this serving configuration; claims about its
sensitivity are explicitly out of scope.
\emph{(v) Directional interpretation.} The toward-actor tendency reported in
\cref{sec:direction} is a qualitative observation; a quantitative directional
attribution requires interaction types where the actor remains ahead of the
ego throughout the horizon and a richer counterfactual design than the present
paired protocol supports.
\emph{(vi) No matched-control interventions.} Steered conditions are compared
only against a zero-bias control, which shows that the trajectory responds to
the bias but not that it responds to the \emph{actor}. Controls holding the
biased token count fixed while varying what those tokens depict (a shifted or
mirrored box, a background region, a second vehicle, a scattered token set)
would separate actor identity from token count and image position. This is the
most important missing experiment, and we claim no actor specificity without
it.

\section{Conclusion}
\label{sec:conclusion}

We studied inference-time attention steering for a VLA autonomous driving
model. A bounded additive pre-softmax bias on Alpamayo-R1's Qwen3-VL backbone,
applied to detector-grounded visual tokens of real vehicles, produces a
monotonic, paired-significant dose--response in the diffusion trajectory
decoder on 50 lane-change scenarios. The effect is layer-localized to late layers, depth-accumulating,
exactly absent at zero bias under paired seeds, and stands out against a
sampler noise floor an order of magnitude larger. Across the scenario set the
steered trajectory tends to shift toward the side of the attended actor, with
the magnitude growing in both bias scale and the number of biased layers;
mapping a desired safety behavior onto bias sign, magnitude, and placement is
the central open design question this framework now enables. A per-call
exposure audit shows that the intervention never reached the autoregressive
reasoning pathway in this stack, reframing the observed CoC invariance as
verified exposure rather than demonstrated robustness. The most immediate next
step is a set of matched-control interventions, which is what would license a
claim of actor specificity rather than sensitivity to visual-attention
perturbation in general. Beyond that, future work will establish (or bound)
language-path exposure, study asymmetric and negative biases for
clearance-increasing behavior, and extend to closed-loop metrics on additional
VLA architectures and interaction types.

%
%
\bibliographystyle{splncs04}
\bibliography{main}
\end{document}